\documentclass[runningheads]{llncs}

\usepackage{eccv}
\usepackage{eccvabbrv}
\usepackage{graphicx}
\usepackage{booktabs}
\usepackage{amsmath,amssymb}
\usepackage{hyperref}

\newcommand{\surf}{\mathrm{surf}}
\newcommand{\parf}{\mathrm{par}}
\begin{document}

\title{Joint-Conditioned Stereo Surface Reasoning for Interaction Field Estimation}
\titlerunning{Joint-Conditioned Stereo Surface Reasoning}
\author{Yanlin Jin\inst{1}\textsuperscript{*} \and
Yifan Yang\inst{2}\textsuperscript{*} \and
Bowen Yang\inst{1}\textsuperscript{*,\textdagger} \and
Kai Zhu\inst{1}}
\authorrunning{Y. Jin et al.}
\institute{Ant Digital Technologies, Ant Group
\and Institute of Automation, Chinese Academy of Sciences (CASIA)\\
\email{ybw377690@antgroup.com}\\
\textsuperscript{*}Equal contribution. \quad
\textsuperscript{\textdagger}Corresponding author.}
\maketitle

\begin{abstract}
Predicting hand--object interaction fields requires locating the nearest object-surface point for each hand joint, often from small and partially occluded image regions. We view this task as joint-conditioned surface-endpoint estimation: each joint has its own nearest endpoint, while endpoints from the same hand can draw on shared local surface evidence. This structure motivates Joint-Conditioned Stereo Surface Reasoning (JSSR). A temporal-stereo network jointly predicts 3D joints, a direct interaction field, and per-view endpoint evidence. Calibrated candidate search evaluates endpoint hypotheses using joint-specific image compatibility and cross-view correspondence. A hand-shared candidate support lets joints draw on common surface evidence, and a learned residual gate controls the geometric correction when observations are ambiguous. Our system built on this method ranked third on the \href{https://hands-workshop.org/challenge2026.html}{SHOW3D Interaction Field Challenge} leaderboard.
\end{abstract}

\section{Introduction}
Interaction fields encode spatial hand--object relations~\cite{Fan_2023_CVPR}. The SHOW3D task~\cite{Rim_2026_CVPR} represents each visible hand by 21 vectors from its joints to the nearest object-surface points. Predicting these vectors requires resolving joint-to-surface associations despite small image regions, mutual occlusion, and depth ambiguity.

Our key observation is that each field endpoint, $\mathbf e_j=\mathbf p_j+\mathbf f_j$, is a projectable location whose image support can be evaluated across calibrated views. Moreover, endpoints associated with one hand can draw on a common set of surface hypotheses. Selection must remain joint-conditioned because different joints have different nearest surface points. This motivates combining stereo correspondence with shared candidate support.

\begin{figure}[tbp!]
\centering
\includegraphics[width=0.96\textwidth]{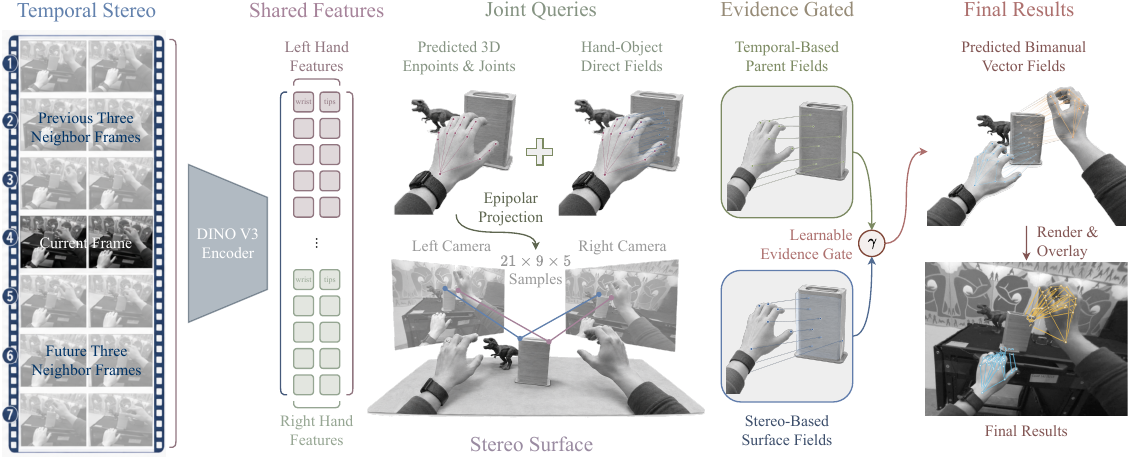}
\caption{\textbf{Method overview.} A temporal-stereo joint-query network predicts 3D joints, a direct interaction field, and per-view endpoint estimates. Calibration forms initial 3D endpoints, which are refined using stereo candidate search and hand-shared surface evidence. An evidence-dependent gate applies the geometric field correction only when reliable. The rightmost panel shows a real prediction.}
\label{fig:overview}
\end{figure}

We propose Joint-Conditioned Stereo Surface Reasoning (JSSR), illustrated in Fig.~\ref{fig:overview}. Our temporal-stereo network jointly predicts 3D joints, a direct field, and per-view endpoint evidence; calibrated search and hand-shared support progressively refine the endpoint estimates. Since image evidence can remain ambiguous, a learned residual gate controls the surface correction. The resulting design uses geometric evidence to guide field prediction while retaining the direct prediction as a fallback.

\section{Method}
\subsection{Network Overview}
The network predicts the interaction field for a target time step from seven synchronized stereo pairs. A shared, trainable DINOv3 encoder~\cite{Simeoni_2025_DINOv3} processes both views, hand-aware temporal aggregation summarizes neighboring observations, and a Transformer decoder assigns one query to each of the 42 joints. Prediction heads estimate the camera-0 joint $\widehat{\mathbf p}_j$ and direct field $\widehat{\mathbf f}^{\rm dir}_j$. For the geometric stage, the same query attends to current-frame tokens to obtain a soft 2D endpoint and a pooled feature from which its depth is predicted. Calibration maps the two view estimates into camera 0 and learned confidences combine them into a coarse endpoint.

\subsection{Joint-Conditioned Stereo Surface Reasoning}
The endpoint formulation is essential for exploiting stereo: unlike a free 3D offset, an endpoint is projectable into both calibrated views and can therefore be verified by cross-view correspondence. We use the predicted camera-0 2D endpoint location to define a calibrated ray and sample $K=32$ candidate depths along it. Let $\mathbf x^{v}_{j,k}$ be the feature sampled at candidate $k$ in view $v$ and $\mathbf q_j$ the joint query. Its evidence score is
\begin{equation}
\begin{aligned}
s_{j,k}&=\frac{1}{2\sqrt d}\!\sum_{v\in\{L,R\}}\!\langle\mathbf q_j,\mathbf x^{v}_{j,k}\rangle
 +\cos(\mathbf x^{L}_{j,k},\mathbf x^{R}_{j,k}),\\
P_{j,k}&=\operatorname{softmax}_k(s_{j,k}),\qquad
\widehat{\mathbf e}^{\rm ray}_j=\sum_{k=1}^{K}P_{j,k}\mathbf c_{j,k}.
\end{aligned}
\label{eq:epipolar}
\end{equation}
where the two terms measure joint-conditioned image compatibility and stereo feature agreement, respectively; invalid projections are masked before normalization. A reliability-weighted D4RT descriptor~\cite{Zhang_2025_D4RT} adds an auxiliary logit in our final implementation, but only reorders these fixed candidates.

Independent rays may still select unrelated surfaces. We therefore expand their local hypotheses and pool them into one hand-level candidate set. Each joint retrieves $\widehat{\mathbf e}^{\rm surf}_j$ from this shared support using stereo appearance and relative 3D geometry, encouraging a coherent surface without requiring a CAD model or object pose. Candidate construction, learned scoring, and supervision are detailed in the supplementary material.

\subsection{Evidence-Gated Field Refinement}
Because stereo matching may fail under occlusion or blur, we retain the direct--epipolar prediction as a stable parent and apply the surface estimate only as a gated correction:
\begin{equation}
\widehat{\mathbf f}^{\rm final}_j=\widehat{\mathbf f}^{\rm par}_j+\gamma_j\left[\left(\widehat{\mathbf e}^{\rm surf}_j-\widehat{\mathbf p}_j\right)-\widehat{\mathbf f}^{\rm par}_j\right].
\label{eq:final}
\end{equation}
The zero-initialized gate $\gamma_j$ predicts how far to move from the parent toward the surface field. It is set to zero for invalid candidates and supervised by the interpolation coefficient that best approaches the ground truth; details are provided in the supplementary material.

\section{Experiments}
\textbf{Protocol.} After removing 535 geometrically inconsistent annotations, we randomly hold out contiguous clips, leaving 72,307 training and 12,201 evaluation targets with no temporal overlap. All DINOv3 variants share the same initialization and train independently for eight epochs at $512\!\times\!512$ using AdamW, backbone/head learning rates of $10^{-5}/10^{-4}$, paired flipping, and global batch 16. We report epoch~8 without model-specific selection.

\textbf{Ablations.} ResNet-50 is a scale reference. \emph{DINO-Direct} regresses the 3D field from the target stereo pair; \emph{Temporal7} adds six neighboring pairs. \emph{Two-view Endpoints} predicts per-view 2D endpoints and depths, maps them to camera~0 by calibration, and confidence-fuses them without epipolar search. \emph{Full JSSR} adds 32-depth epipolar search, hand-shared surface selection, and gated fusion.

\begin{table}[ht!]
\centering
\caption{Fixed epoch-8 clip-held-out results and historical challenge scores. ADE is in mm; Acc@10 is in \%.}
\label{tab:block-ablation}
\scriptsize
\setlength{\tabcolsep}{4pt}
\begin{tabular}{lrrrrr}
\toprule
Method & Left & Right & Mean ADE & Acc@10 & LB score$\downarrow$ \\
\midrule
Official ResNet-50 & 29.256 & 29.088 & 29.172 & 26.91 & 58.00 \\
DINO-Direct & 15.995 & 16.562 & 16.278 & 50.04 & 41.61 \\
\quad + Temporal7 & 15.809 & 16.578 & 16.194 & 49.65 & 37.08 \\
\quad + Two-view Endpoints & 15.720 & 16.130 & 15.925 & 50.59 & 36.51 \\
\quad + Full JSSR & \textbf{15.520} & \textbf{15.961} & \textbf{15.740} & \textbf{51.23} & \textbf{33.59} \\
\bottomrule
\end{tabular}
\par\smallskip
\parbox{\linewidth}{\scriptsize\textit{Note.} LB scores match historical submissions by method family, with different training settings; they are not scores of the epoch-8 ablation checkpoints. The Temporal7 submission also uses D4RT, and the endpoint submission includes epipolar search.}
\end{table}

\noindent\textbf{Results.} Temporal7, Two-view Endpoints, and Full JSSR successively reduce mean ADE to 16.194, 15.925, and 15.740~mm. The complete stereo reasoning improves the temporal baseline by 0.454~mm (2.80\%).

\noindent\textbf{Challenge result.} Our best single-model submission scored 33.59. Object-specific fine-tuning and a fixed 50/50 ensemble of two expert systems yielded our best leaderboard score of 32.61 (third place).

\bibliographystyle{splncs04}
\bibliography{references}

\clearpage
\appendix
\pdfbookmark[0]{Supplementary Material: Learned Shared-Surface Reasoning}{supplementary}
\begin{center}
{\Large\bfseries Supplementary Material:\\[0.3em]
Learned Shared-Surface Reasoning}
\end{center}
\medskip
\section{Shared Local Surface Support}

The epipolar stage estimates one parent endpoint
$\bar{\mathbf e}_{h,j}=\widehat{\mathbf p}_{h,j}+\widehat{\mathbf f}^{\parf}_{h,j}$
for joint $j$ of hand $h$. Selecting every joint independently from its own ray can nevertheless produce endpoints on mutually inconsistent structures. We therefore construct a single local 3D candidate support for each hand and let all 21 joint queries select from it. This is a geometry-structured \emph{soft} surface constraint: it does not reconstruct a mesh and does not use a CAD model or object pose.

\subsection{Candidate construction}

We project each of the 21 parent endpoints into camera 0. Around each projected location, we use five image-plane offsets (center, left, right, up, and down) and nine log-spaced depth multipliers in $[\exp(-0.25),\exp(0.25)]$. Back-projecting these hypotheses produces
\begin{equation}
\mathcal C_h=\left\{\mathbf c_{h,j,p,\ell}\right\},\qquad
|\mathcal C_h|=21\times5\times9=945.
\label{eq:support}
\end{equation}
The spatial radius is adapted to the projected endpoint spread and clamped to $[0.04,0.08]$ in normalized image coordinates. Importantly, the joint index is discarded after construction: every query may select any candidate in the hand-level set $\mathcal C_h$. Each candidate is transformed and projected into both calibrated views. Candidates with non-finite coordinates, non-positive depth, or projections outside either valid view are masked.

\subsection{Learned stereo candidate score}

For candidate $k$, let $\mathbf x_k^L$ and $\mathbf x_k^R$ denote features sampled from the two current-frame feature maps. We combine stereo appearance with explicit projective geometry:
\begin{align}
\mathbf g_k &={\left[\mathbf u_k^L,\;\mathbf u_k^R-\mathbf u_k^L,\;
\log(z_k/z_{\rm base})\right]},\\
\mathbf z_k &=\phi\!\left(\left[\tfrac12(\mathbf x_k^L+\mathbf x_k^R),\;
|\mathbf x_k^L-\mathbf x_k^R|,\;\mathbf x_k^L\odot\mathbf x_k^R,\;\mathbf g_k\right]\right),
\label{eq:pair-feature}
\end{align}
where $\phi$ is an MLP and $z_{\rm base}$ is the median parent-endpoint depth for the hand. The mean, absolute difference, and elementwise product expose complementary stereo matching cues; $\mathbf g_k$ supplies image position, disparity, and relative depth.

The decoded joint representation is mapped to a query $\mathbf q_{h,j}$. The logit for assigning candidate $k$ to joint $j$ is
\begin{equation}
a_{h,j,k}=\frac{\langle\mathbf q_{h,j},\mathbf z_{h,k}\rangle}{\sqrt{48}}
+r(\mathbf z_{h,k}),
\label{eq:surface-score}
\end{equation}
where the first term is joint-specific compatibility and the learned scalar $r(\mathbf z_{h,k})$ measures candidate quality shared across joints. After masking invalid candidates,
\begin{equation}
P_{h,j,k}=\operatorname{softmax}_k(a_{h,j,k}),\qquad
\widehat{\mathbf e}^{\surf}_{h,j}=\sum_kP_{h,j,k}\mathbf c_{h,k}.
\label{eq:surface-endpoint}
\end{equation}
Thus, calibration determines where evidence may be sampled, while the MLP learns how stereo appearance and projective geometry indicate a plausible local surface point.

\section{Training the Surface Distribution and Gate}

For a candidate endpoint $\mathbf c_{h,k}$, its field relative to the predicted joint is
\begin{equation}
\mathbf f^{(k)}_{h,j}=1000\,\mathbf c_{h,k}-\widehat{\mathbf p}_{h,j}
\quad\text{(mm)}.
\end{equation}
We turn distances to the ground-truth field into a soft target distribution,
\begin{equation}
\pi_{h,j,k}\propto
\exp\!\left[-\frac{\|\mathbf f^{(k)}_{h,j}-\mathbf f^{\rm gt}_{h,j}\|_2^2}
{2\sigma^2}\right],\qquad \sigma=35\ \text{mm},
\label{eq:soft-target}
\end{equation}
and minimize $D_{\rm KL}(\pi\|P)$. Both the field error and $\sigma$ are expressed in millimeters; $35$~mm is equivalent to $0.035$~m. An endpoint loss additionally supervises the expected surface field
$\widehat{\mathbf f}^{\surf}_{h,j}=1000\widehat{\mathbf e}^{\surf}_{h,j}-\widehat{\mathbf p}_{h,j}$.

The surface estimate is applied as a residual correction rather than replacing the parent prediction:
\begin{equation}
\widehat{\mathbf f}^{\rm final}_{h,j}=\widehat{\mathbf f}^{\parf}_{h,j}
+\gamma_{h,j}\left(\widehat{\mathbf f}^{\surf}_{h,j}-\widehat{\mathbf f}^{\parf}_{h,j}\right).
\label{eq:gate}
\end{equation}
The gate is a zero-initialized MLP with a $\tanh$ output. It receives the joint representation, pooled surface feature, selection entropy, and expected candidate score, and is forced to zero when no valid candidate exists. To supervise it, define $\mathbf d=\widehat{\mathbf f}^{\surf}-\widehat{\mathbf f}^{\parf}$ and the best convex interpolation coefficient
\begin{equation}
\alpha=\operatorname{clip}_{[0,1]}
\frac{(\mathbf f^{\rm gt}-\widehat{\mathbf f}^{\parf})^\top\mathbf d}
{\|\mathbf d\|_2^2+\epsilon}.
\label{eq:gate-target}
\end{equation}
The gate is trained with a Smooth-$L_1$ loss toward $\alpha$. The complete additional objective is
\begin{equation}
\mathcal L_{\surf}=0.02\,D_{\rm KL}(\pi\|P)
+0.05\,\frac{\operatorname{ADE}(\widehat{\mathbf f}^{\surf},\mathbf f^{\rm gt})}{100}
+0.05\,\operatorname{SmoothL1}(\gamma,\alpha).
\label{eq:surface-loss}
\end{equation}

\enlargethispage{7\baselineskip}
\paragraph{Interpretation.}
The shared support couples the 21 predictions because their endpoint distributions are defined over the same candidate set. It does not force different joints to choose the same 3D point; instead, it restricts them to hypotheses supported by one common local volume and a common stereo scoring function. Consequently, ``surface'' denotes shared, stereo-validated endpoint support rather than an explicitly reconstructed object surface.

\end{document}